\documentclass[10pt,twocolumn,letterpaper]{article}

\usepackage{cvpr}
\usepackage{times}
\usepackage{epsfig}
\usepackage{graphicx}
\usepackage{amsmath}
\usepackage{amssymb}

\usepackage{times}
\usepackage{epsfig}
\usepackage{graphicx}
\usepackage{amsmath}
\usepackage{amssymb}

\usepackage{kotex}
\usepackage{xcolor}
\usepackage{ctable}

\usepackage{multirow}

\usepackage{threeparttable}

\usepackage{makecell}

\usepackage[breaklinks=true,bookmarks=false]{hyperref}

\cvprfinalcopy 

\def\cvprPaperID{****} 

\begin{document}

\title{M2FN: Multi-step Modality Fusion for Advertisement Image Assessment}

\author{Kyung-Wha Park\\
Interdisciplinary Program in Neuroscience, Seoul National University.\\
{\tt\small kwpark@bi.snu.ac.kr}
\and
Jung-Woo Ha\\
NAVER AI LAB, NAVER CLOVA.\\
{\tt\small jungwoo.ha@navercorp.com}
\\\\
JungHoon Lee\\
Statistics and Actuarial Science, Soongsil University.\\
{\tt\small ssutartup@gmail.com}
\and
Sunyoung Kwon\\
School of Biomedical Convergence Engineering, Pusan National University.\\
{\tt\small skwon@pusan.ac.kr}
\and
Kyung-Min Kim\\
NAVER AI LAB, NAVER CLOVA.\\
{\tt\small kyungmin.kim.ml@navercorp.com}
\and
Byoung-Tak Zhang\\
Interdisciplinary Program in Neuroscience, Seoul National University.\\
Department of Computer Science and Engineering, Seoul National University.\\
Surromind Robotics.\\
{\tt\small btzhang@bi.snu.ac.kr}
}

\maketitle

\begin{abstract}
Assessing advertisements, specifically on the basis of user preferences and ad quality, is crucial to the marketing industry. Although recent studies have attempted to use deep neural networks for this purpose, these studies have not utilized image-related auxiliary attributes, which include embedded text frequently found in ad images. We, therefore, investigated the influence of these attributes on ad image preferences. First, we analyzed large-scale real-world ad log data and, based on our findings, proposed a novel multi-step modality fusion network (M2FN) that determines advertising images likely to appeal to user preferences. Our method utilizes auxiliary attributes through multiple steps in the network, which include conditional batch normalization-based low-level fusion and attention-based high-level fusion. We verified M2FN on the AVA dataset, which is widely used for aesthetic image assessment, and then demonstrated that M2FN can achieve state-of-the-art performance in preference prediction using a real-world ad dataset with rich auxiliary attributes. 
\end{abstract}

\section{Introduction}
{\let\thefootnote\relax\footnote{{\textbf{This paper has been published in Applied Soft Computing.}\\ANOVA~:~analysis of variance; AVA~:~aesthetic visual analysis; BERT~:~bidirectional encoder representations from transformers; CBN~:~conditional batch normalization; CTR~:~click-through rate; EMD~:~earth mover's distance; KLD~:~Kullback--Leibler divergence; LCC~:~linear correlation coefficient; LLF~:~low-level fusion; LSTM~:~long short-term memory; M2FN~:~multi-step modality fusion network; MCD~:~minimum covariance determinant; MLP~:~multilayer perceptron; NIMA~:~neural image assessment; OCR~:~optical character recognition; SPRC~:~Spearman rank correlation; SOTA~:~state-of-the-art; VQA~:~visual question answering}}} 
Unlike other industries of the 20th century, the advertising industry has rapidly expanded only in recent times. The global advertising market, already worth 532.5 billion dollars in 2019, is expected to reach 769.9 billion dollars by 2024\footnote{https://www.imarcgroup.com/global-advertising-market}. Of this large market, nearly one third is based on online advertising, which has achieved enormous commercial success in the modern world. The market value of online advertising is projected to grow significantly from 193 billion dollars in 2017 to 236 billion dollars by 2026. 

Given the immense size and scope of the modern advertising industry, much competition comes from within the industry. Businesses and advertisers, therefore, must ensure that their advertisements (hereinafter referred to as ads) are more attractive and compelling than each other’s.

A well-created ad not only attracts user attention but also leads to engagement. From the perspective of online advertisers, an effective ad would not only attract more interest but also generate more clicks, leading to increased product purchases with fewer ad displays, thereby reducing the cost of bidding on the ad schedule. Therefore, a precise quantitative evaluation of ads is important for advertisers and ad designers.

Accordingly, this study conducts an ad image assessment that involves understanding image preferences: which images are preferred and to what extent is an image preferred over others. In addition, an initial study is provided via a deep network visualization: of which part of the image might be preferred.

Existing image assessment methods typically use an image as the only input. Some studies~\cite{talebi2018nima, lee2019image} have proposed models to quantitatively evaluate human-subjective aesthetic judgments. Additionally, various studies have been conducted in the fields of neuroscience, cognitive science, computational photography, psychology, marketing, and aesthetics to assess the emotional impressions and aesthetic qualities of images~\cite{talebi2018nima,6247954,daugherty2016research}.

However, predicting image preferences for ads differs from conventional aesthetic-image assessment in two ways. First, previous aesthetic-image assessment studies~\cite{talebi2018nima, lee2019image} did not utilize auxiliary attributes such as image annotations, time, and age of participants. Thus, they were ineffective at evaluating ad images because user preferences may depend significantly on various conditions.

Second, unlike aesthetic-image assessment studies that use ratings as a metric, studies on ad image preference use the click-through rate (CTR), the ratio of the number of users who click on the ad to the number of its displays. CTRs are commonly used to measure the success of online advertising; additionally, they can be used to assess image preferences for ads. Although CTRs are not automatically indicative of user preference, they do indicate the extent to which an ad accomplishes its objective (i.e., capturing user attention and engagement), which are induced by user preferences. Most CTR prediction studies on ad image preference rely on large-scale user data~\cite{chen2016deep}, which are sensitive and expensive to be collected. Therefore, recent studies have focused on reducing dependence on user information and leveraging other attributes.

Despite the differences between aesthetic-image assessment and CTR prediction studies, multi-modal fusion, which integrates information from different inputs, has become essential for both fields to expand the application domain and enhance the performance of deep network models. However, existing CTR prediction studies on ad images that apply multi-modal fusion do not utilize image-driven auxiliary attributes, but only facilitate linguistic attributes (e.g., ad title and description) and mainly involve simple operations, such as concatenation~\cite{chen2016deep,xia2019deep}. 

Handling complex ad images using simple concatenation-based modality fusion is difficult. Therefore, we extensively analyzed large-scale, real-world, and in-house log data from online game ads, including the CTR for exposure events, to utilize auxiliary attributes more effectively in assessing ad images. We determined that auxiliary attributes interact with the visual information of ad images at different conceptual levels, depending on each attribute. For example, at a low level, the overall color of an ad image could depend on the season. At a high level, the position or content of the embedded text could vary depending on the users' ages. We required techniques that intervene in the early vision process and integrate auxiliary attributes with the refined vision representation in the later stages.

Owing to the recent advent of vision and language fusion studies, two techniques exist that match our needs---conditional batch normalization (CBN)~\cite{de2017modulating} and attention mechanism. The CBN technique modulates the visual processing in the early stage based on other input modalities. The attention mechanism is an essential technique in recent multi-modal deep learning frameworks~\cite{attendxu15, kim2018multimodal}. Both these techniques are mostly used for fusing language modalities into visual representations. We experimentally extended their scope for visual processing by leveraging auxiliary attributes, which consist of content-related attributes such as linguistic attributes (e.g., title and dominant colors) and classical metadata (e.g., age and time).

It is not straightforward to explain user preferences toward ad images using a single pipeline of a deep neural network, because such an architecture would contain an excessive number of layers to analyze. Inspired by this finding, we proposed a multi-step modality fusion network (M2FN) that incorporates auxiliary attributes, such as user-related metadata and visual--linguistic features, to assess ad images. To integrate ad images and auxiliary attributes effectively, we performed multiple steps of modality fusion and employed CBN~\cite{de2017modulating} and a spatial attention mechanism for predicting CTR. To evaluate our approach and the performance of our model, we validated M2FN on two datasets: the aesthetic visual analysis (AVA) dataset~\cite{6247954}, which is an aesthetic image assessment dataset, to test the quality of image assessments made by M2FN; and a dataset of real-world ad images, to evaluate the model using CTR predictions. We achieved state-of-the-art (SOTA) results on both datasets. In addition, we used a visualization method for each step in M2FN to empirically determine the image regions that influence user preferences as a initial study.

The main contributions of our research are summarized below:
\begin{itemize}
\item We propose a novel deep neural networks model for advertisement image assessment that involves performing multiple steps of modality fusion via conditional batch normalization and a spatial attention mechanism.
\item We leverage visual--linguistic attributes for both aesthetic and advertisement image assessment tasks, along with classic metadata, denoted as auxiliary attributes.
\item We demonstrate state-of-the-art performance for both tasks.
\item We show the contribution of this model with ablation studies and visualizations to investigate the effectiveness of each step and auxiliary attribute.
\end{itemize}

\section{Related Research}
\textbf{Image Assessment: }Image assessment conventionally refers to the evaluation of an image in terms of quality (such as enhancement and noise). Visual aesthetic assessment, a form of image assessment, focuses on art, beauty, and personal preference. Driven by the importance of aesthetic assessment, the AVA dataset, which is a large-scale image database accompanied by various metadata and rich annotations, was introduced for advanced research~\cite{6247954}. Using the AVA dataset, several studies have performed aesthetic assessments of images by learning universal traits that characterize aesthetic quality~\cite{talebi2018nima,deng2017image,yu2018aesthetic}. Although aesthetic assessment generally depends on individual tastes, some universal rules, such as the golden ratio, color harmonies, and rule of thirds~\cite{datta2006studying,dhar2011high,ke2006design,luo2011content,luo2008photo}, have been reported. To obtain a higher correlation with human ratings, a study~\cite{talebi2018nima} focused on distributions of ratings instead of aggregated mean scores using the earth mover's distance (EMD)~\cite{zhang2018unreasonable}. Most visual aesthetic assessment studies concentrate on image content and do not involve additional textual metadata. \\\\
\textbf{Vision and Language: }Research on visual--linguistic representation learning has become popular since the advent of visual question answering (VQA) challenges~\cite{antol2015vqa}.
Generally, learning methods comprise two key components:
\begin{enumerate}
\item Attention mechanism: Specific parameters are used to train models to attend to particular regions of images or correlated words in sentences~\cite{attendxu15}. Because of the sparsity of regions that carry useful information in images or videos, most methods reduce the redundancy on spatio-temporal inputs based on other modalities, e.g., text~\cite{kim2018multimodal}. It typically finalizes the fusing step using matrix multiplication methods between other modalities.
\item Multi-modal fusion: The fusion of multi-modal data, which provides information characterized by different properties or formats for computational models, can lead to enhanced performance. While two distinct levels of fusion, i.e., early and late, were widely used~\cite{PORIA201798}, recent computational models can handle various multi-modal representations; moreover, the levels of fusion have become more diverse. Various methods utilizing multi-modal representations, ranging from simple methods, such as concatenation or element-wise addition, to complex methods, such as compact bilinear pooling~\cite{fukui-etal-2016-multimodal}, have been proposed. 
\end{enumerate}
CBN is a fusion method that initially modulates vision by controlling the batch normalization parameters in layers of a visual representation model based on other inputs~\cite{de2017modulating}. It trains multilayer perceptrons (MLPs), which predict the mean and variance for in-depth normalization. 
Our approach adopts ideas from learning methods of visual--linguistic representations. We utilized auxiliary attributes via multi-step fusion based on both CBN and attention-based fusion mechanisms. In recent studies, methodologies for effectively performing bi-modal fusion of vision and language representations have been developed~\cite{nguyen2018improved,Perez-Rua_2019_CVPR}. These methodologies search for hidden representations or structures to achieve the best performance within multi-layer networks. While these methods can be applied to bi-modality problems, the problem in this study involves more than two modes. Through extensive experiments, we attempted to determine the optimal output from layers via multi-modal fusion. \\\\
\textbf{Advertisement Preference Prediction: }
Recently, because of the success of deep learning, neural network-based approaches for predicting CTRs have been proposed. Automatic and flexible feature learning methods (directly from raw images) have been introduced~\cite{mo2015image} as an alternative to conventional, custom-developed, feature-based approaches. For more accurate predictions, Mo et al.~\cite{mo2015image} used ad information, such as detailed item categories and display positions, in addition to raw images, and achieved improved performance. Other CTR prediction methods involve building end-to-end deep learning architectures that learn representative features~\cite{chen2016deep,xia2019deep}. These models are trained using both raw images and other related information, such as ad zones, ad groups, item categories, and user features. Whereas these approaches use additional meta-information together with images, their use of simple concatenation-based modality fusion leaves room for improvement. End-to-end architectures and single pipeline designs also make it challenging to determine the reason for the superiority of an ad image.

\section{Preliminaries}
In this section, we define the terms relevant to our study. First, \textbf{impression} is a common term that represents the number of \textbf{exposures} of an ad to users. After an exposure, \textbf{clicks} can be collected from users, which are assumed to be induced by their \textbf{preference} or the attractiveness of the images; note, however, that the number of clicks is not automatically indicative of user preferences. In this paper, we use the term ``preference" to describe both attractiveness and user preference. Furthermore, the term \textbf{ad insights} refers to observed phenomena that are discovered through statistical analysis of advertising data. Examples of ad insights are presented in the sections on datasets.

{
\setlength{\tabcolsep}{10pt}
\ctable[star,
    caption = {Auxiliary attributes from the Real-Ad dataset.},
    label = {table:detailed_items_dataset},
    pos = th,
    center,
    doinside = \footnotesize
]{clll}{}{
    \toprule
\multicolumn{2}{c}{Feature name} & \makecell[c]{Description} & \makecell[c]{Format} \\ 
               \midrule\midrule
 \multirowcell{12}{Classical \\ metadata} & Gender      &  Gender of user   & \{1(female), 2(male)\}      \\
 \cmidrule{2-4}
                            & Age         & Age of user   & \{1(under 14), 2(15--19), …, 9(over 50)\}       \\
                            \cmidrule{2-4}
                            & Month & Month    & \{1(Jan), 2(Feb), …, 12(Dec)\}      \\ 
                            \cmidrule{2-4}
                            & Weekday & Weekday   & \{1(Mon), 2(Tue), …, 7(Sun)\}       \\ 
                            \cmidrule{2-4}
                            & Time & Time   & \{0, 1, 2, …, 23\}       \\
                            \cmidrule{2-4}
                            & Position & Position of ad in platform app & \{1, 2, 3, 4\} \\ 
                            \cmidrule{2-4}
                            & Cate2 & Mid-category of item & \makecell[l]{\{1(casual), 2(gambling),\\ 3(mid-core), 4(core)\}}\\
                            \cmidrule{2-4}
                            & Cate3 & Low-category of item & \makecell[l]{\{1(RPG), 2(sports), 3(simulation),\\ 4(action), 5(adventure), 6(strategy),\\ 7(casino), 8(casual), 9(puzzle)\}}\\
                           
                      \midrule\midrule
   \multirowcell{8}{Content \\ attributes} 
                           & Title         & Title of ad, embedded using BERT   & 768-dimensional embedding vectors\\
                           \cmidrule{2-4}
                           & Desc & \makecell[l]{Description of ad, embedded using BERT} & 768-dimensional embedding vectors \\
                           \cmidrule{2-4}
                           & OCR & \makecell[l]{Detected from image using open API,\\ embedded using BERT} & 768-dimensional embedding vectors \\ 
                           \cmidrule{2-4}
                           & Domcol          & \makecell[l]{Dominant color of image labeled using \\ K-means-based algorithm}   & \makecell[l]{\{1(black), 2(blue), 3(brown), 4(green), \\5(grey), 6(multiple), 7(pink), 8(red),\\ 9(white), 10(yellow)\}}      \\

    \bottomrule
}
}

\section{Auxiliary Attributes}
\begin{figure}[t]
		\includegraphics[width=\columnwidth]{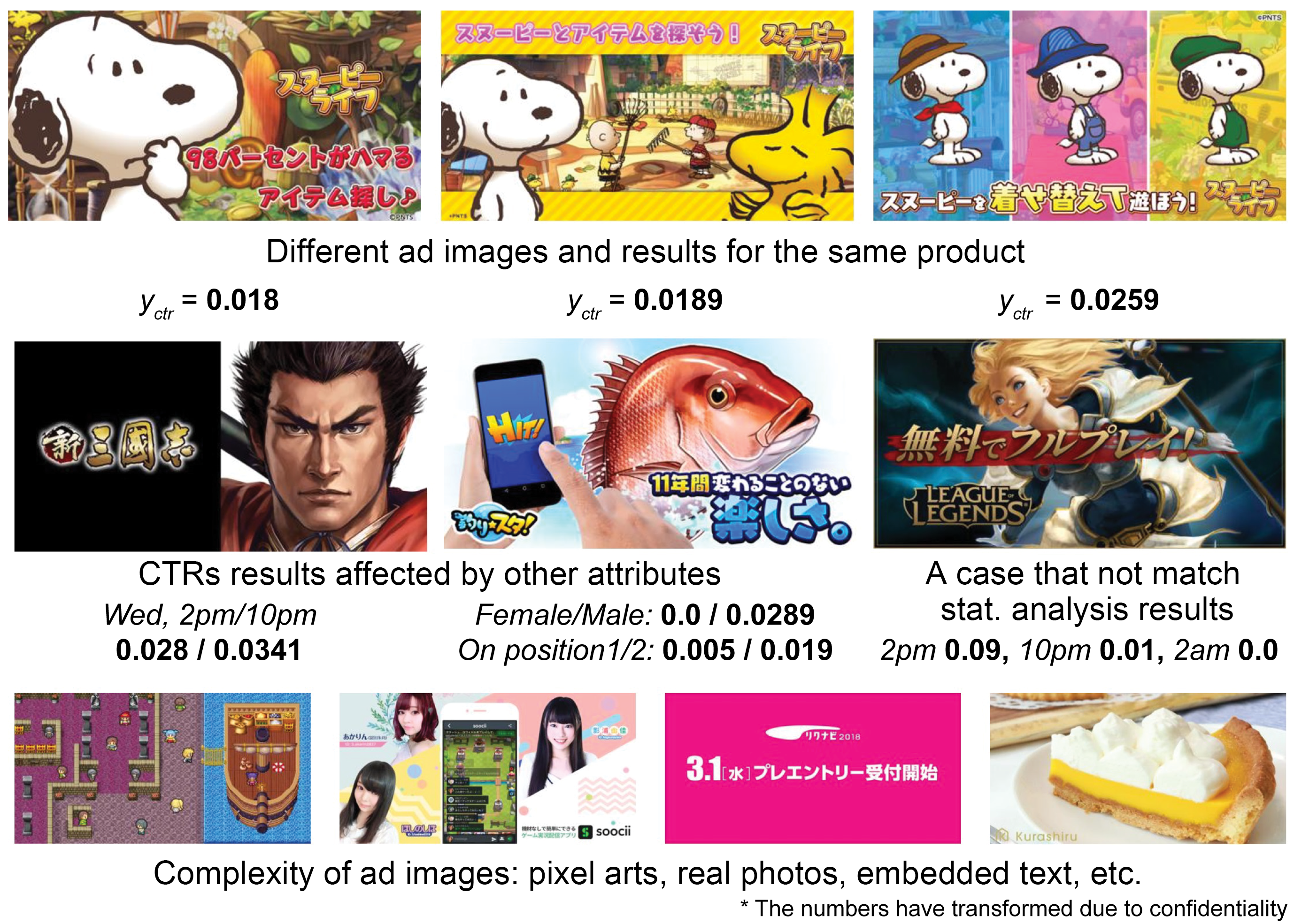}
		\caption{Examples of advertisement (ad) images and CTR prediction results.}
		\label{examples_of_ad}
\end{figure}
Auxiliary attributes denote a superset of classical metadata, where
metadata is a common term for data that describe a user or item. Currently, such information used in most studies comprises the personal data of users, which are sensitive and expensive to collect. We use the expression ``auxiliary attributes'' to describe expanded metadata that describe the content, which, in this study, is an ad image. We exploited several features, including visual--linguistic attributes, which are outlined in \tablename~\ref{table:detailed_items_dataset}. In contrast to 
most existing image assessment studies \cite{talebi2018nima, lee2019image} that focus only on images for preference assessment, we used additional features as well. In several studies investigating CTR, the input data were full of metadata, such as user demographical data (e.g., gender and age) and ad-exposure event-related data (e.g., weekday and time). 
However, an ad image assessment model should be capable of using rich auxiliary attributes related to images, such as brands and embedded text~\cite{mitchell1986effect}, and colors~\cite{mehta2009blue,labrecque2012exciting}, including classic user-related metadata and ad display policy. Auxiliary attributes are crucial for assessing ad images. For example, the top row in Fig.~\ref{examples_of_ad} shows different ad images for the same product. Among these images, users prefer the image on the right, which uses embedded text (e.g., ``please change the character's outfit'') to request for their participation. Other attributes, such as the time of ad display, which are listed in the middle row, may also affect user engagement. Some cases may even be contrary to the overall statistical analysis results, as shown in the middle row's right. For further demonstration, various ad images consisting of pixel arts, real photos, and embedded text are shown at the bottom row in Fig.~\ref{examples_of_ad}.
In this study, all these metadata were used together with additional data from various sources, e.g., embedded text, which is expected to express the intention of an ad. To allow computational models to learn these catchphrases, visual--linguistic attributes were used in the Real-Ad dataset. In addition, the dataset possesses an optical character recognition (OCR) attribute with titles and descriptions, which are common linguistic attributes in image datasets. Details on the utilized auxiliary attributes are explained in the datasets section. 

\section{M2FN: Multi-step Modality Fusion Network}
\begin{figure*}[t]
	\begin{center}
	\includegraphics[width=\textwidth]{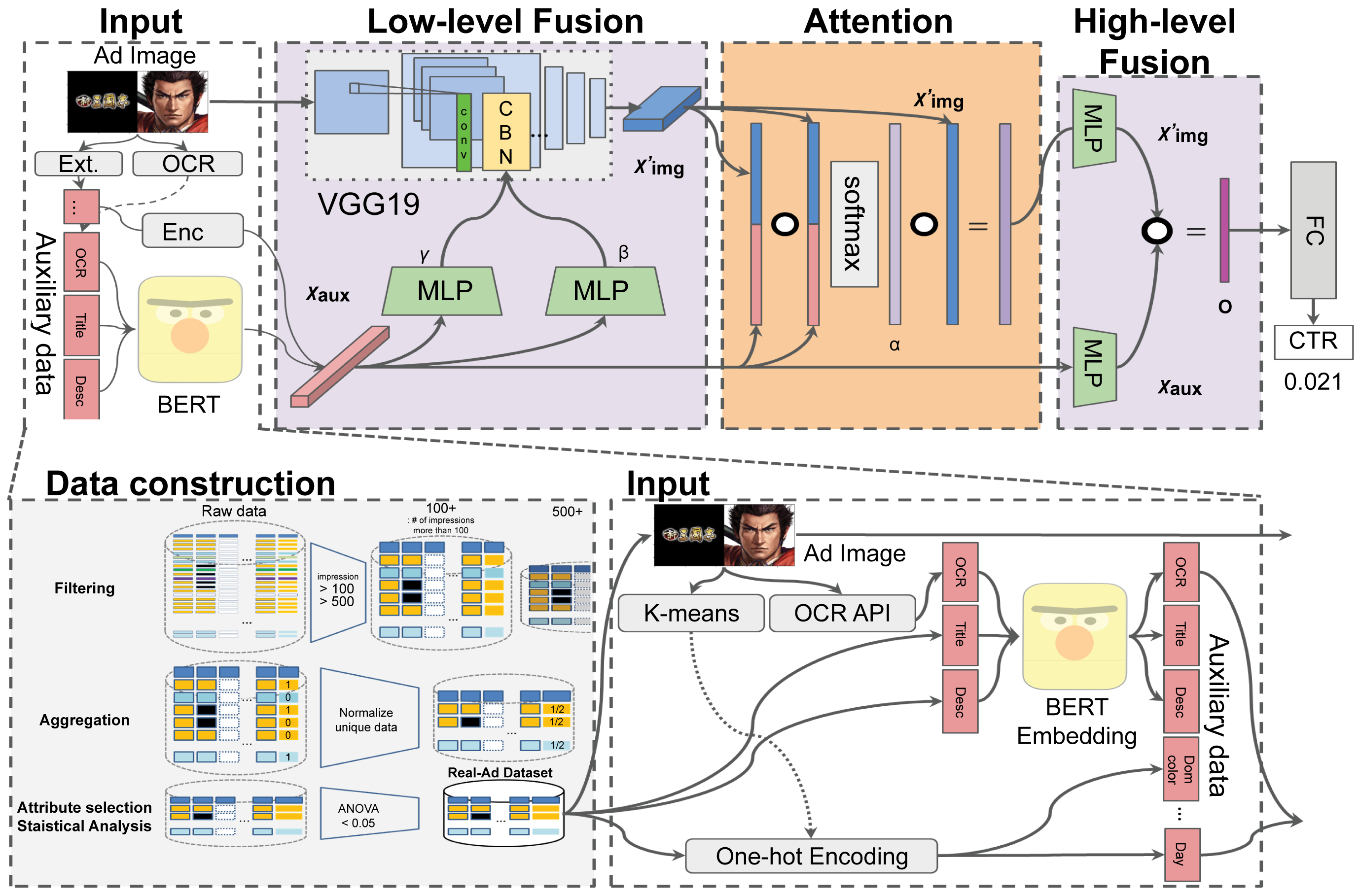}
		\caption{Overall structure from data input phase to multi-step modality fusion network (M2FN).}
		\label{overall_structure}
	\end{center}
	\vspace{-3mm}
\end{figure*}

M2FN consists of four main modality fusion steps as depicted in Fig.~\ref{overall_structure}: i) input step, which integrates features using the conventional early-fusion method, wherein auxiliary attributes are concatenated to be processed in the next step; ii) low-level fusion step, which deals with specific features, such as dominant colors of ad images and auxiliary attributes; iii) spatial attention step, which addresses text-expression location in ad images; and iv) high-level fusion step, which involves abstracted visual features and auxiliary attributes, such as demographics, time, and semantics of linguistic information.

\subsection{Low-Level Fusion Step}
We hypothesized that by providing each layer of the network a hint of the complexity of the ad image, auxiliary attributes can improve prediction performance. To achieve this, we employed CBN~\cite{de2017modulating} for low-level fusion. CBN modulates the layers of the network instead of performing simple concatenation. Additionally, it trains shallow neural networks that predict the scale factor parameter $\gamma$ and shift factor parameter $\beta$ in batch normalization from the input data. In VQA, for example, the question sentences are embedded using long short-term memory (LSTM), and the question embedding $e_q$ is fed to the CBN module. In this study, instead of text questions, auxiliary attributes are used as input to the CBN. Before being input to the CBN, categorical attributes are encoded into a one-hot vector, whereas text information is represented as a bidirectional encoder representations from transformers (BERT)-based embedding vector.

Similar to batch normalization, CBN can be performed at any layer of a neural network. In this study, CBN was used to combine images and auxiliary attributes only after the first convolution of the early stage VGG-19, based on SOTA results.

\subsection{Spatial Attention Step}
The location of textual information and its expression in ad images have a crucial influence on the CTR of an ad. To address this, we introduced an attention mechanism that considers spatial relationships for modality fusion. 

More specifically, image features and auxiliary embedding vectors were represented using $N_b \times C \times W \times H$ and $N_b \times  dim_{aux}$ tensors, respectively, where $N_b$, $C$, $W$, and $H$ denote batch size, number of channels, width, and height, respectively. The auxiliary embedding vectors were replicated to be a $N_b \times  dim_{aux} \times W \times H$ tensor. The auxiliary tensor was concatenated with the image feature tensor to be a $N_b \times (C+dim_{aux}) \times W \times H$ tensor. Thereafter, it was fed to a fully connected layer of an MLP. The resulting vectors $\{N_b, W \times H\}$ obtained using the SoftMax function became the attention matrix. This attention matrix was multiplied (Hadamard product) with image features to achieve a soft-attention map.

\subsection{High-Level Fusion Step}
The role of high-level fusion is to emphasize the effect of spatial relationships between visual features and auxiliary attributes by integrating two modalities near the output layer. Unlike in previous approaches, where a simple concatenation was used, a matrix element-wise multiplication between the output of the attention mechanism and auxiliary attributes was performed, as in~\cite{de2017modulating}. To match the dimension size, affine transformation was performed using linear layers for each image feature and auxiliary attributes. The activation function of the hyperbolic tangent was then applied to both outputs, and the resulting vectors were multiplied element-wise. There was a significant difference in the regression performance with and without this fusion mechanism, which is detailed in the experiments section.

\subsection{Loss Function}
Our model uses the impression-weighted mean squared error as a loss function:
\begin{equation}
L = \frac{1}{N}\sum\nolimits_{n = 1}^N {{w_n} \cdot {{\left( {{{\hat y}_n} - {y_n}} \right)}^2}},
\end{equation}
\noindent where $w_n$, ${\hat y}_n$, and ${y_n}$ denote the impressed number, predicted CTR, and real CTR, respectively, of the $n$-th instance, and $N$ represents the data size.

In addition, Kullback--Leibler divergence (KLD) was used as a loss function when the preference score of the data was represented as a distribution form similar to that of the AVA dataset~\cite{6247954}:
\begin{equation}
{L_{KLD}} = \frac{1}{N}\sum\nolimits_{n = 1}^N {p({y_n}) \cdot \log \frac{{p({y_n})}}{{p({{\hat y}_n})}}},
\end{equation}
\noindent where $p({y_n})$ and $p({{\hat y}_n})$ refer to the distributions of the real CTR and the CTR predicted via SoftMax function, respectively, of the $n$-th instance.

\section{Datasets}
\begin{figure}[]
		\includegraphics[width=\columnwidth]{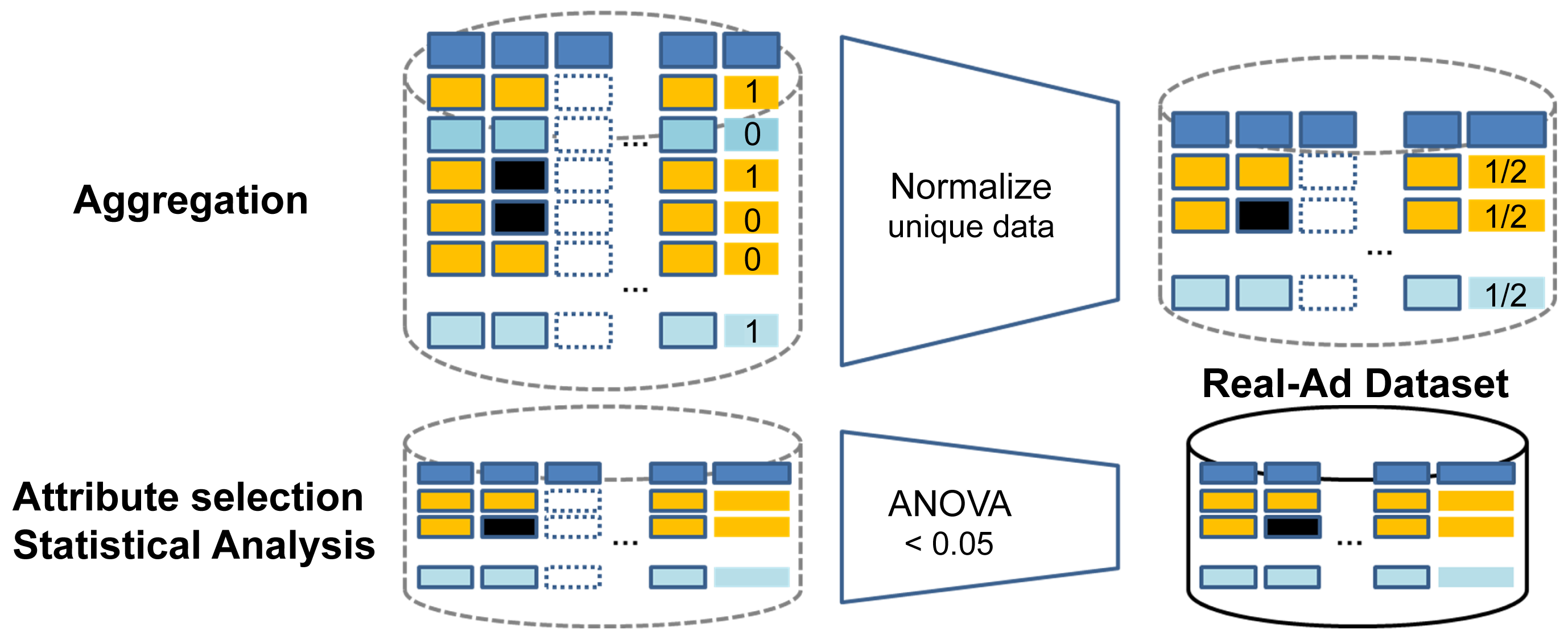}
		\caption{Schematic showing aggregation and attribute selection in data construction. Rightmost column in each bin diagram contains the label.}
		\label{data_preparation_diagram}
\end{figure}

\subsection{Real-Ad Dataset} 
For evaluating the model, we used a large-scale dataset of click logs of ad image impressions, collected from a global online ad service in Japan in 2018. The total number of impressions is approximately 500 million and includes 3,747 distinct ad images, as shown in \tablename~\ref{table:stat_real_ad_dataset}. While the training dataset was being constructed, aggregation and attribute selection were performed on the data. These processes are illustrated in Fig~\ref{data_preparation_diagram}. Note that this dataset cannot be released because of confidentiality issues.

\subsection{Data Construction}
Before the dataset was constructed, irrelevant logs were discarded to mediate the survivorship bias~\cite{mangel1984abraham}; instances of malicious users were removed before the ad banner was read.

{
\setlength{\tabcolsep}{5pt}
\ctable[
    caption = {Details on Real-Ad dataset. Each instance in raw dataset represents one exposure of an ad. Aggregated 100+ and 500+ datasets were constructed from over 100 and 500 impressions, respectively.
    },
    label = {table:stat_real_ad_dataset},
    pos = h,
    doinside = \footnotesize
]{ccc rrr}{}{
    \toprule
    \multicolumn{3}{c}{Dataset} & \# instances & \# clicks & \# ad images  \\ 
    \midrule\midrule
    Raw(Game)&& & 500M & 20M & 3,747 \\ 
    \midrule
    \multirow{4}{*}{Aggregated}&\multirow{2}{*}{100+}&Train & 353,510 & 800K & 3,045 \\
    &&Test & 173,248 & 330K & 1,436 \\
    \cmidrule{2-6}
    &\multirow{2}{*}{500+}&Train & 47,325 & 350K & 1,519 \\
    &&Test & 24,003 & 140K &  681 \\
    \bottomrule
}
}
\vspace{-2mm}

\tablename~\ref{table:stat_real_ad_dataset} outlines  the specifications of the Real-Ad dataset. One instance of raw data corresponds to one exposure event of an image ad with auxiliary attributes. Each instance has a label of 1 or 0, denoting whether it was clicked or not, respectively. Each ad was viewed multiple times by different users. Moreover, each ad was displayed at different times, ad positions, etc.

Each instance was aggregated to CTR: 
\begin{equation}
{y_n} = \frac{{M_n^c}}{{{M_n}}},    
\end{equation}
\noindent where $M_n^c$ denotes the number of clicks, and $M_n$ denotes impressions for the $n$-th unique instance.

The aggregated data were compressed and compared with the raw data, although these aggregated data remained suitable for ad image assessment. In the raw data, more than 99\% of the impressions were not clicked, leading to severe imbalance problems. Because of this imbalance, label conflicts and errors were frequent at an instance level. In contrast, aggregated CTR values were represented as real numbers with a higher resolution, which facilitated comparative analysis with existing image assessment studies~\cite{talebi2018nima}.

We separated the training and test datasets. Every experimental result was obtained from unseen image data from the test set.

\subsection{Statistical Analysis of Real-Ad Data: Ad Insights}
\begin{figure*}[t]
    \begin{center}
		\includegraphics[width=0.95\textwidth]{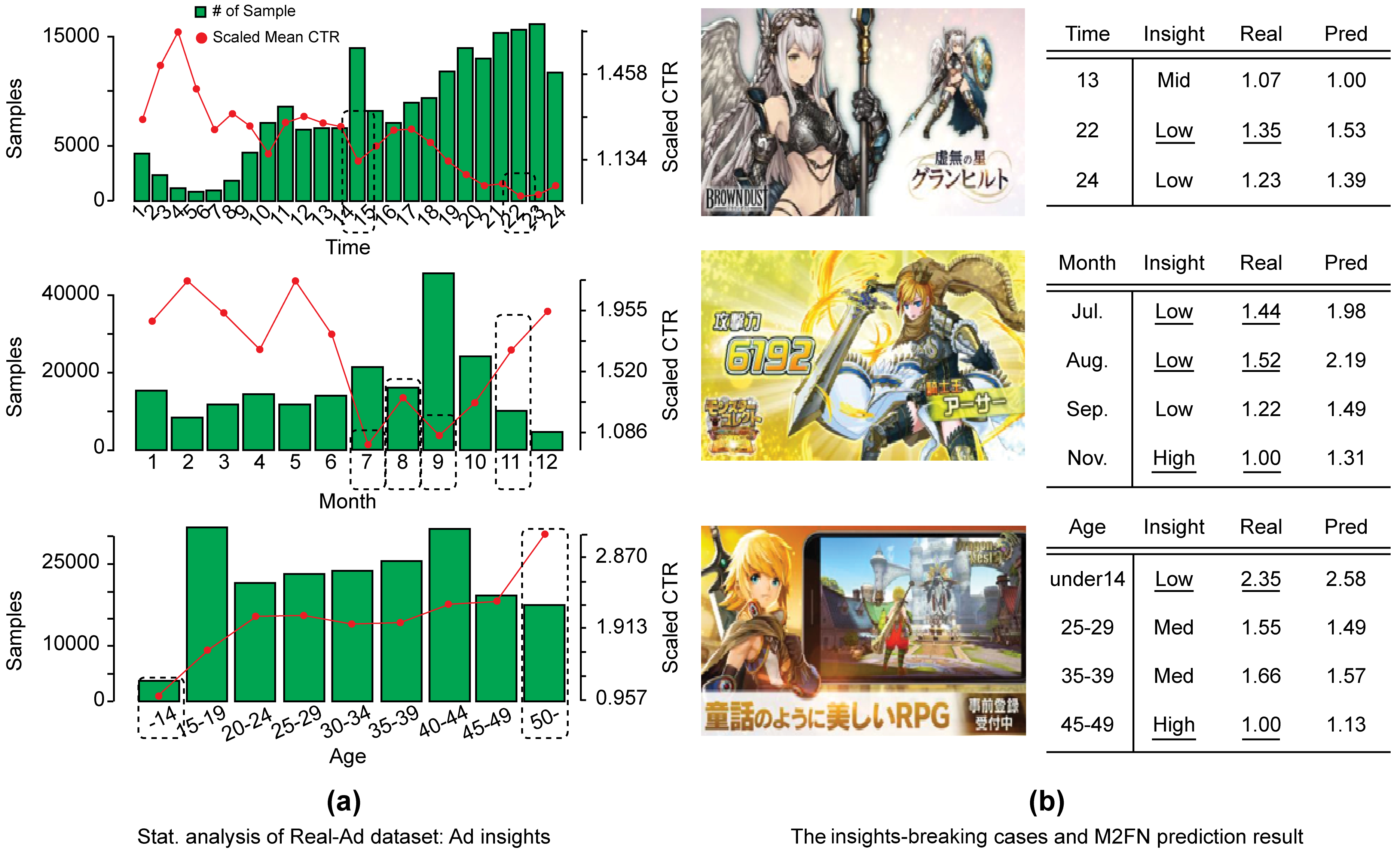}
		\caption{Ad insights and task introduction. (a) Ad insights and statistical analysis results for Real-Ad. (b) Counter-evidence of insights derived from (a) and CTR prediction results of M2FN. Although table contents shown in (b) differ from results in (a), M2FN exhibits reliable performance. CTR values are divided by the smallest CTR in each table because of confidentiality issues.}
		\label{myth_counter_cases}
	\end{center}
	\vspace{-3mm}
\end{figure*}

Through statistical analysis methods such as analysis of variance (ANOVA) and logistic regression analysis, we investigated the influence of ad images and their auxiliary attributes on CTR. Through statistical analyses, we identified the relationship between specific CTR responses and each auxiliary attribute. Generally, these characteristic movements (ad insights) are strategically considered points identified by marketers to increase the CTRs of ads. Fig.~\ref{myth_counter_cases}(a) illustrates that the ad insights of three of these CTR-effective attributes usually affect the ad performance regardless of the category of the item. For example, the age bar plot in Fig.~\ref{myth_counter_cases} shows that game ads have a higher probability of being clicked when they are shown to elder users. In addition, we reaffirmed that the dominant color\footnote{http://info.rocketfuel.com/} and visual--linguistic attributes also significantly influence the effectiveness of ads~\cite{mitchell1986effect}. 

Furthermore, the time and month bar plots in Fig.~\ref{myth_counter_cases}(a) show the influence of time-sequential attributes on CTR. In detail, ads published at mid-dawn (2 AM to 4 AM), mid-morning (9 AM to 11 AM), noontime (12 PM), and mid-afternoon (3 PM to 5 PM) exhibited better results than those published at other hours. Additionally, the beginning and end of a week exhibited higher CTRs than the other days. Finally, the lowest CTRs occurred during the third quarter of the year, i.e., July, August, and September.

In contrast, many cases exist that do not follow these ad insights. Fig.~\ref{myth_counter_cases}(b) shows the unusual cases discovered during our statistical analyses of Real-Ad, and the results of CTR prediction by our model. These results demonstrate that preference prediction on ad images is highly challenging. For example, CTR distributions for the same ad image and auxiliary attributes varied depending on different time attributes, which can be considered a covariate shift.

Our statistical analyses provided some interesting observations. In most images, the brand logo was observed in a variety of ways. The logo shape is known to have a considerable influence because it conveys the image of the brand~\cite{walsh2006consumer}. Additionally, there were specific dominant colors (black, blue, brown, etc.) that correlated with higher CTR instances. Designing to represent even low-level visual parts is therefore important. According to these insights, a sophisticated model design must leverage auxiliary attributes and integrate them with ad images effectively.

\subsubsection{Attribute Selection}
The raw Real-Ad dataset contains approximately 40 attributes, which were collected with click logs. Some of these attributes are redundant or weakly related to CTR. Attribute selection was therefore performed to cleanse the dataset and select only the auxiliary attributes; it was not performed on the images. To select attributes that are crucial for CTR prediction, we employed ANOVA and logistic regression analysis at a significance level of 0.05. Through these analyses, we selected nine key attributes responsible for CTR. These attributes are outlined in \tablename~\ref{table:detailed_items_dataset}.

\subsubsection{Attribute Preprocessing}
The dominant color and visual--linguistic auxiliary attributes had to be preprocessed for computational models to be trained. These auxiliary attributes included title, description, and embedded textual expression of ad images (OCR). The dominant color of an ad image was represented as an element of a predefined color set, which included 10 colors. We used $K$-means clustering with a minimum covariance determinant (MCD) distance to extract the intermediate dominant color from each image. After the intermediate color was obtained, it was mapped to one of the predefined dominant colors.

In addition, we supplemented the data with visual--linguistic auxiliary attributes, such as title, description, and OCR result. The title and description were collected from the ad application form. OCR results were obtained using an open-source API~\cite{baek2019character,baek2019STRcomparisons} to identify the letters in the ad images. This API detects scene text by exploring each character and affinities between characters. Thus, it can flexibly detect complicated scene text images, such as curved or deformed texts, which frequently appear in ad images. Unique sentences were collected for each visual--linguistic auxiliary attribute, with 1,583,\hspace{0.3em} 2,695, and 2,695 sentences being collected for title, description, and OCR, respectively. We embedded these sentences into vectors using BERT~\cite{Wolf2019HuggingFacesTS,devlin-etal-2019-bert}. Other non-linguistic auxiliary attributes were represented via one-hot encoding.

While grouping and summing CTRs, we considered only the instances with more than 100 and 500 impressions for the feasibility assessment of CTR values. The levels of attributes that have less than 50,000 impressions were merged with the closest level to cope with biases in the dataset. The total number of dimensions of auxiliary attributes was 2,383, of which 2,304 were visual--linguistic attributes (with 768 dimensions for each).

\subsection{Benchmark Dataset: AVA Dataset}
M2FN can contribute to a conventional image assessment task. We evaluated our model on AVA~\cite{6247954}, which is an image dataset designed for aesthetic assessment studies. This dataset comprises images and their annotations, including aesthetic, semantic, and photographic-style annotations. The photographic-style annotations are composed of 14 styles (complementary colors, duotones, HDR, etc.) that represent the camera settings. However, these annotations were not used in this study because an excessive number of cases exist without annotations.

Semantic annotations are textual tag data, e.g., \textit{nature, black and white, landscape, animal}. There are approximately 200,000 images, and each image has at least one tag. The tag set size is 67 when no-tag cases for image-only CTR predictions are added. Each text tag was represented as a BERT embedding vector. In this study, embedding vectors with 768 dimensions were used to represent visual--linguistic auxiliary attributes.

Aesthetic annotations refer to the rating data per image from hundreds of amateur and professional photographers. These data were used as score labels and characterized by a histogram distribution ranging from 1 to 10. In this study, the dataset was divided into training and test sets, with an 8:2 ratio for benchmark experiments.

\section{Experimental Results} 
\subsection{Experimental Setup}
\textbf{Evaluation Metrics:} Depending on the dataset, the resulting outputs were in distribution (10 buckets) or scalar value (score) form. Performance evaluation was based on ranking. Spearman rank correlation (SPRC) and the linear correlation coefficient (LCC) were computed to rank the output scores and compare them with the ground-truth ranking. If the output score was in the form of a distribution, the SPRC and LCC were calculated and examined for both the mean and standard deviations. \\

{
\setlength{\tabcolsep}{10pt}
\ctable[star,
    caption = {Shape and number of training parameters of each layer in M2FN.
    },
    label = {table:num_parameters},
    pos = t,
    center,
    doinside = \footnotesize
]{ccc rr}{}{
    \toprule
    \multicolumn{2}{c}{Layer name} & Symbol & \multicolumn{1}{c}{Shape} & \# parameters \\ 
    \midrule\midrule
    \multirow{2}{*}{Input} & & $x_{img}$ & $224 \times 224 \times 3$  & \\ 
    & & $x_{aux}$ & $2385 \times 1$ & \\ 
    \midrule\midrule
    \multirow{4}{*}{Low-level fusion}&\multirow{2}{*}{CBN}& $\gamma$ &  $x_{aux} \times 512 + 512 \times \textbf{256}$ & 1.3M\\
    && $\beta$ &  $x_{aux} \times 512 + 512 \times \textbf{256}$ & 1.3M \\
    \cmidrule{2-5}
    & VGG19 & - & - & 14.7M \\
    & w/o classif. layers &$x'_{img}$ & $7 \times 7 \times 512$ & \\
    \cmidrule{1-5}
    \multirow{2}{*}{Attention}& Stack & $[x'_{img} + x_{aux}]$ & $[25088 + 2385]$ & \\
    & Embedding &&  $stack \times \textbf{512} + \textbf{512} \times 1$ & 14M \\
    & Alpha & $\alpha$ &  $7 \times 7 \times 512$ &  \\
    & Soft attention & $x'_{img} \odot \alpha$ &  $7 \times 7 \times 512$ & \\
    \cmidrule{1-5}
    \multirow{3}{*}{High-level fusion}& Embedding Soft att & $x''_{img}$ & $25088 \times \textbf{512}$ & 13M\\
    & Embedding $x_{aux}$ & $x'_{aux}$ &  $2385 \times \textbf{512}$ & 1.2M \\
    & Fusion & $x''_{img} \times x'_{aux}$ &  $512 \times 512$ &  \\
    \cmidrule{1-5}
    \multirow{2}{*}{Output}& FC layers & & $512 \times 4096 + 4096 \times 4096 $ & 19M\\
    & FC layer out & & $4096 \times 1$ & 4k \\
    
    \bottomrule
}
}

\noindent \textbf{Model Parameter Settings:} The shape and the number of training parameters for each layer in M2FN are outlined in Table~\ref{table:num_parameters}. Because VGG19 is widely used and well-known, the table contents are omitted, and only the total number of parameters is presented. 

The benchmark and Real-Ad datasets have distinct hyperparameter settings for M2FN. Among the four major modules, CBN (low-level fusion), attention, and high-level fusion should determine the hidden layer size of the MLP for embedding representations inside the module. When the AVA datasets, which have relatively small numbers of auxiliary attributes, were trained, the hyperparameters of the modules were 64, 512, and 512, respectively. When the Real-Ad dataset was used, the hyperparameters were set to 256, 512, and 512. The hyperparameters are highlighted in bold in the third column of Table~\ref{table:num_parameters}. This decision was based on over 20,000 automated parallel hyperparameter-searching experiments.

The batch size of the dataset was 128, and five P40 GPUs were used for training. All models were trained for 100 epochs. 

All the experiments are implemented and performed based on NAVER Smart Machine Learning (NSML) platform~\cite{kim2018nsml,sung2017nsml} with AutoML~~\cite{kim2018chopt,park2019visualhypertuner}.

\subsection{Comparison of Prediction Performance}
We compared our M2FN with a simple concatenation-based modality fusion model and neural image assessment (NIMA)~\cite{talebi2018nima}, which is an SOTA method for image assessments, as the baselines. Table~\ref{table:model_comparison_real_ad} compares their CTR prediction performance on the Real-Ad dataset. Unlike AVA, which includes rating counts in the form of ten buckets for each image, Real-Ad provides only a mean of the CTRs. Because our proposed method requires training on a CTR value, the CTR distributions were approximated based on a log--normal distribution that resembles CTRs. When converted to this distribution, it is represented as ``dist''; Otherwise, it is represented in Table~\ref{table:model_comparison_real_ad} as ``regr.'' Previously, the best performing model (NIMA) on the benchmark dataset used EMD as a loss function. The asterisk (*) signifies that, instead of EMD, Kullback–Leibler divergence (KLD) was applied as a loss function (for ``regr.'' cases, the impression-weighted MSE was used, as mentioned in a previous section). Regardless of the number of impressions (100+ or 500+), our method clearly performed better. For this dataset, KLD, compared with EMD, appears to be a better choice for the loss function.

Table~\ref{table:model_comparision_AVA} compares our models with baseline models using the AVA dataset to verify that our model can be effective at image assessments, with a quantitative comparison between NIMA (InceptionV2)~\cite{deng2017image}, the previous SOTA model, and M2FN. 

{
\setlength{\tabcolsep}{4pt}
\ctable[
    caption = {Performance comparison on the Real-Ad dataset. In the header row, m and std. represent mean and standard deviation, respectively. Dist. and Regr. denote distribution and regression, respectively.},
    label = {table:model_comparison_real_ad},
    pos = h,
    doinside = \footnotesize
]{cclccccc}{}{
    \toprule
\multicolumn{3}{c}{Models}                      & SPRC(m) & LCC(m) & SPRC(std.) & LCC(std.) \\ 
               \midrule\midrule
\multirow{5}{*}{100+} & \multirow{3}{*}{Dist.}   & NIMA          & 0.110   & 0.121  & 0.146      & 0.142     \\
                      &                           & NIMA*         & 0.289   & 0.290  & 0.139      & 0.143     \\
                      &                           & \textbf{M2FN} & 0.344   & 0.367  & 0.172      & 0.175     \\ 
                      \cmidrule{2-7}
                      & \multirow{2}{*}{Regr.} & NIMA          & 0.325   & 0.343  & -          & -         \\
                      &                           & \textbf{M2FN} & \textbf{0.384}   & \textbf{0.381}  & -          & -         \\ 
                      \midrule
\multirow{5}{*}{500+} & \multirow{3}{*}{Dist.}   & NIMA          & 0.308   & 0.249  & 0.165      & 0.166     \\
                      &                           & NIMA*         & 0.453   & 0.448  & 0.190      & 0.190     \\
                      &                           & \textbf{M2FN} & 0.484   & 0.478  & 0.216      & 0.190     \\
                      \cmidrule{2-7}
                      & \multirow{2}{*}{Regr.}  & NIMA          & 0.501   & 0.451  & -          & -         \\
                      &                           & \textbf{M2FN} & \textbf{0.561}   & \textbf{0.530}   & -          & -         \\
    \bottomrule
}
}

{
\setlength{\tabcolsep}{15pt}
\ctable[star,
    caption = {Performance comparison on AVA dataset. In the header row, m and std. represent mean and standard deviation, respectively. ``-cat'' denotes simple concatenation of both auxiliary and image, which represents models without low-level fusion in existing studies. ``-LLF'' signifies low-level fusion. If there is no additional explanation and table entry is denoted by ``-,'' the model was trained only on images. M1FN is a prototype model of M2FN. Numbers in brackets refer to blocks wherein low-level fusion is either conducted (1) or not conducted (0).},
    label = {table:model_comparision_AVA},
    doinside = \footnotesize,
    center
]{lccccc}{}{
    \toprule
    \multicolumn{1}{c}{Models} & \# parameters & SPRC(m) & LCC(m) & SPRC(std.) & LCC(std.) \\ 
    \midrule\midrule
    NIMA(InceptionV2) &20M& 0.612 & 0.636 & 0.218 & 0.233 \\
    NIMA(VGG16) &134M& 0.592 & 0.610 & 0.202 & 0.205 \\
    \midrule
    InceptionV4 &40M& 0.606 & 0.619 & 0.183 & 0.198 \\
    InceptionV4-cat &40M& 0.576 & 0.581 & 0.160 & 0.155 \\
    \midrule
    Resnet101 &44M& 0.496 & 0.470 & - & - \\
    Resnet101-cat &44M& 0.549 & 0.560 & 0.192 & 0.189 \\
    Resnet101-LLF &47M& 0.558 & 0.561 & 0.233 & 0.230 \\
    Resnet50-LLF &28M& 0.544 & 0.552 & 0.225 & 0.223 \\
    \midrule\midrule
    VGG19 &139M& 0.617 & 0.622 & - & - \\ 
    M1FN(VGG19-cat) &139M& 0.572 & 0.581 & 0.198 & 0.197 \\ \addlinespace
    \midrule
    M2FN \{1,1,1,1,1\} &78M& 0.552 & 0.553 & 0.183 & 0.176 \\ 
    M2FN \{0,0,0,0,1\} &65M& 0.572 & 0.579 & 0.200 & 0.198 \\ 
    M2FN \{0,0,1,1,1\} &72M& 0.575 & 0.573 & 0.203 & 0.196 \\ 
    M2FN \{0,0,1,0,0\} &65M& 0.590 & 0.601 & 0.230 & 0.229 \\ 
    M2FN \{1,1,1,0,0\} &72M& 0.612 & 0.613 & 0.253 & 0.266 \\ 
    \midrule
    \textbf{M2FN \{1,0,0,0,0\}}&65M& \textbf{0.630} & \textbf{0.640} & \textbf{0.310} & \textbf{0.322} \\
    \bottomrule
}
}

The results of the comparison based on the necessity and location of multi-modal fusion are also listed in the table. ``-cat'' denotes simple concatenation of both the auxiliary and image, whereas ``-LLF'' represents low-level fusion. If a data entry in the table is denoted by ``-,'' the model was trained on images only (no auxiliary attributes).

In \tablename~\ref{table:model_comparision_AVA}, the number of training parameters are included to show the complexity of each model. Experiments with InceptionV4, a deeper version of the previous SOTA, confirmed that deepening the network does not always improve the task performance. This inference can also be verified using the results for ``Resnet101,'' which is widely used for image-related tasks~\cite{he2016deep}. Although it has numerous layers, the model is not reliable. From the ``-cat'' results for ``InceptionV4-cat'' and ``VGG19-cat,'' simple concatenation was inferred to be degrading the performance, compared with the corresponding image-only cases (``InceptionV4'' and ``VGG19''). In contrast, in the case of ``Resnet101-cat,'' simple concatenation enhanced the performance. Because the visual feature extraction module (``Resnet101'') is large, it overfits easily and severely degrades performance. Enhanced performance is assumed to have been achieved primarily with auxiliary attributes. ``VGG19'' exhibited better performance compared with the overall results, and the best performance among image-only cases. Furthermore, because we reduced the sizes of fully connected layers at the end of M2FN via hyperparameter searching, M2FN is less complex than VGG19, which is partially nested in M2FN for visual feature extraction.

Furthermore, we conducted other comparison experiments to assess how the location of low-level fusion affects performance. Because VGG19 consists of five blocks, the five numbers in the parentheses depict the location of the fusion. If fusion is performed inside the block, it is written as ``1.'' 

The results demonstrated better performances when the fusion was located closer to the input stage. Existing studies apply CBN to all blocks~\cite{de2017modulating}; however, this adversely affects the task performance. 

As mentioned in the introduction section, ad images have a complex structure that differs from those of ordinary images. These results support our intuition behind the multi-step structure, based on the ad insights and previous studies. M2FN, which uses multi-step modality fusion, performed best on the benchmark dataset.

{
\setlength{\tabcolsep}{6pt}
\ctable[
    caption = {M2FN ablation study results on AVA.},
    label = {table:ablation_module_AVA},
    pos = h,
    doinside = \footnotesize
]{cccc|cc|cc}{}{
\toprule
\multicolumn{4}{c}{Module}      & \multicolumn{2}{c}{Mean} & \multicolumn{2}{c}{Std.} \\ 
\midrule
Aux&Low&Att&High                         & SPRC             & LCC                              & SPRC                             & LCC \\ 
\midrule\midrule
$\times$&$\times$&$\times$&$\times$          & 0.584              &   0.592                      &  0.214                             &  0.231\\
\midrule
O&$\times$&$\times$&$\times$              & 0.572    &  0.567          & 0.194                &  0.194\\ 
O&O&$\times$&$\times$               & {0.618}           &   {0.586}      &  {0.266}                &  {0.293}\\
\midrule
O&$\times$&O&$\times$               & 0.565                    &  0.561                          & 0.197                        &  0.201\\
O&$\times$&$\times$&O               & 0.573                 &  0.557                          & 0.217                          &  0.2\\
O&O&O&$\times$                     & 0.608                &   0.601                        &  0.23                            &  0.296\\
O&O&$\times$&O                    & 0.601                & 0.611                          &  0.251                           &  0.285\\ 
\midrule
\multicolumn{4}{c|}{NIMA(Inception-v2)}          & 0.612              &   0.636                      &  0.218                             &  0.233\\
\midrule
O&O&O&O                         & \textbf{0.630}         & \textbf{0.640}               & \textbf{0.310}                 &   \textbf{0.322}\\
\bottomrule
}
}

{
\setlength{\tabcolsep}{5pt}
\ctable[
    caption = {M2FN ablation study results on Real-Ad.},
    label = {table:ablation_module_real_ad},
    pos = h,
    doinside = \footnotesize
]{cccc|cc|cc}{}{
\toprule
\multicolumn{4}{c}{Module}      & \multicolumn{2}{c}{500+} & \multicolumn{2}{c}{100+} \\ 
\midrule
Aux&Low&Att&High                         & SPRC             & LCC                              & SPRC                             & LCC \\ 
\midrule\midrule
Only&\multicolumn{3}{l|}{Linear regression} & 0.239       & 0.195               & 0.166                 &   0.156\\
Only&\multicolumn{3}{l|}{SVM regression (RBF)} & 0.102         & {0.160}               & {0.183}                 &   {0.126}\\
Only&\multicolumn{3}{l|}{SVM regression (poly)} & {0.242}         & {0.177}               & {0.201}                 &   {0.197}\\
Only&\multicolumn{3}{l|}{MLP} & {0.241}         & {0.193}               & {0.240}                 &   {0.217}\\
\midrule\midrule
Only&\multicolumn{3}{l|}{M2FN w/o image}                         & {0.306}         & {0.295}               & {0.244}                 &   {0.230}\\
\midrule
$\times$&$\times$&$\times$&$\times$          & 0.456              &   0.435                      &  0.315                             &  0.342\\
\midrule
O&$\times$&$\times$&$\times$              & 0.437    &  0.412          & 0.297                &  0.320\\ 
O&O&$\times$&$\times$               & {0.496}           &   {0.498}      &  {0.367}                &  {0.371}\\
\midrule
O&$\times$&O&$\times$               & 0.450                    &  0.463                          & 0.341                        &  0.325\\
O&$\times$&$\times$&O               & 0.475                 &  0.467                          & 0.321                          &  0.293\\
O&O&O&$\times$                     & 0.506                &   0.480                        &  0.356                            &  0.334\\
O&O&$\times$&O                    & 0.554                & 0.528                          &  0.361                           &  0.333\\ 
\midrule
O&O&O&O                         & \textbf{0.561}         & \textbf{0.530}               & \textbf{0.384}                 &   \textbf{0.381}\\

\bottomrule
}
}

\subsection{Ablation Study}
\tablename~\ref{table:ablation_module_AVA} and  \tablename~\ref{table:ablation_module_real_ad} compare the performance changes depending on the presence or absence of four major modules for each dataset. The four primary modules include auxiliary attributes in the input step, low-level fusion, attention, and high-level fusion. In the four consecutive columns, ``O'' denotes activating, and ``$\times$'' signifies deactivating. ``$\times \times \times \times $'' is a model trained using only images on vanilla VGG-19, the base of M2FN. In the case of ``O $\times\times\times$,'' where the auxiliary attributes are activated but the CBN is deactivated, the image and auxiliary attributes are learned through concatenation before the last fully connected layer.

The results in Table~\ref{table:ablation_module_real_ad} demonstrate how essential images are to the ad image assessment task. The results of the four baseline models and M2FN were achieved using only auxiliary attributes without ad images. For the M2FN comparative result, we conducted experiments on blank images by filling the input with zeros. Better results than those of the four baseline models were exhibited, although M2FN has many parameters. Furthermore, a large difference in terms of performance between baseline models and M2FN was observed. In particular, there was a difference of approximately 0.15 compared with the result when only the image was considered (``$\times \times \times \times$'' result, the sixth row in Table~\ref{table:ablation_module_real_ad}).

According to the results in the seventh and eighth rows indicating the presence and absence of low-level fusion, CBN enhances performance. The ninth row of the results also demonstrates the strength of M2FN. This row indicates the presence of high-level fusion. The best performance was achieved when all four modules were combined. The integrated results of the four modules proved that our approach is suitable for evaluating impressions. In Table~\ref{table:ablation_module_AVA}, the row of NIMA data represents existing SOTA performance on AVA. Below the NIMA row are data on the performance of M2FN; M2FN outperforms the previous record.

{
\begin{table}[]
\setlength{\tabcolsep}{6pt}
\centering 
\caption{M2FN ablation study on auxiliary attributes using Real-Ad. Names of each attribute, starting with uppercase letters, represent use of multiple attributes.}
\begin{threeparttable}
\begin{tabular}{c ccc cc c c c}
    \toprule
    &\multicolumn{3}{c}{Used auxiliary attributes}                &&   \multicolumn{1}{c}{SPRC}  &&  \multicolumn{1}{c}{LCC}&\\  
    \midrule\midrule
    &\multicolumn{1}{l}{None}&&                             &&   \multicolumn{1}{c}{0.456}       &&   \multicolumn{1}{c}{0.435}&\\ 
    \midrule
    &\multicolumn{1}{l}{ALL\tnote{1}}&&                             &&   \multicolumn{1}{c}{\textbf{0.561}}       &&   \multicolumn{1}{c}{\textbf{0.530}}&\\ 
    &&\multicolumn{1}{l}{$-$ weekday}   &&&   \multicolumn{1}{c}{\underline{0.454}}       &&   \multicolumn{1}{c}{\underline{0.413}}&\\  
    &&\multicolumn{1}{l}{$-$ description}   &&&   \multicolumn{1}{c}{\underline{0.467}}       &&   \multicolumn{1}{c}{\underline{0.453}}&\\  
    &&\multicolumn{1}{l}{$-$ color}   &&&   \multicolumn{1}{c}{\underline{0.503}}       &&   \multicolumn{1}{c}{\underline{0.470}}&\\  
    &&\multicolumn{1}{l}{$-$ (User+Text)\tnote{2}}   &&&   \multicolumn{1}{c}{0.513}        &&   \multicolumn{1}{c}{0.507}&\\  
    &&\multicolumn{1}{l}{$-$ ocr}   &&&   \multicolumn{1}{c}{0.515}       &&   \multicolumn{1}{c}{0.491}&\\  
    &&\multicolumn{1}{l}{$-$ position}   &&&   \multicolumn{1}{c}{0.516}       &&   \multicolumn{1}{c}{0.487}&\\  
    &&\multicolumn{1}{l}{$-$ time}   &&&   \multicolumn{1}{c}{0.531}       &&   \multicolumn{1}{c}{0.501}&\\  
    &&\multicolumn{1}{l}{$-$ gender} &&&   \multicolumn{1}{c}{0.532}       &&   \multicolumn{1}{c}{0.470}&\\  
    &&\multicolumn{1}{l}{$-$ age}   &&&   \multicolumn{1}{c}{0.535}       &&   \multicolumn{1}{c}{0.486}&\\  
    &&\multicolumn{1}{l}{$-$ User\tnote{3}}  &&&   \multicolumn{1}{c}{0.539}       &&   \multicolumn{1}{c}{0.514}&\\  
    
    \midrule\midrule
    &\multicolumn{1}{l}{Text}&&               &&   \multicolumn{1}{c}{0.422}       &&   \multicolumn{1}{c}{0.369}&\\
    &&\multicolumn{1}{l}{$-$ (title $+$ ocr)}&&                       &   \multicolumn{1}{c}{0.453}       &&   \multicolumn{1}{c}{0.411}&\\
    \midrule
    &\multicolumn{1}{l}{User}&&  &&   \multicolumn{1}{c}{0.437}       &&   \multicolumn{1}{c}{0.397}\\
    \bottomrule
\end{tabular}

\begin{tablenotes}\footnotesize
    \item[1] ALL aux: Text, User, month, weekday, time, position, categories, color
    \item[2] remove Text (title, desc, ocr) with User(gender, age) auxes.
    \item[3] remove User (gender, age) from ALL auxes.
    
\end{tablenotes}
\end{threeparttable}

\label{table:ablation_aux_real_ad}
\end{table}
}

The results in Table~\ref{table:ablation_aux_real_ad} are of an ablation study comparing performance changes with and without auxiliary attributes in the Real-Ad dataset. We speculate that content-related attributes (weekday, time, position, categories, and color) are more informative than other attributes, e.g., user-related attributes, for evaluating impressions. We expected better performance to be achieved even if we use only content-related attributes among auxiliary attributes. The results prove our hypothesis.

In Table~\ref{table:ablation_aux_real_ad}, the performance is sorted in ascending order. If an attribute is excluded and the performance drop is significant, it indicates that the attribute is essential. Therefore, we observe the following: i) The time-sequential attribute (weekday) is vital. ii) According to the bottom of the table (-(title + ocr)), the description has negligible effect when used alone. However, description has a significant impact when it is integrated with other attributes. iii) As previous studies have asserted, the color attribute proved essential for advertising content in the table. iv) The user attribute is not crucial, even though it has been extensively used in previous studies. v) However, when the user attribute is combined with others (``text aux'' in the table), performance degradation seems to be substantial. User attribute is assumed to enhance performance when integrated with other attributes.

\subsection{Advertisement Image-Assessment Results}

\begin{figure*}[]
    \begin{center}
		\includegraphics[width=0.95\textwidth]{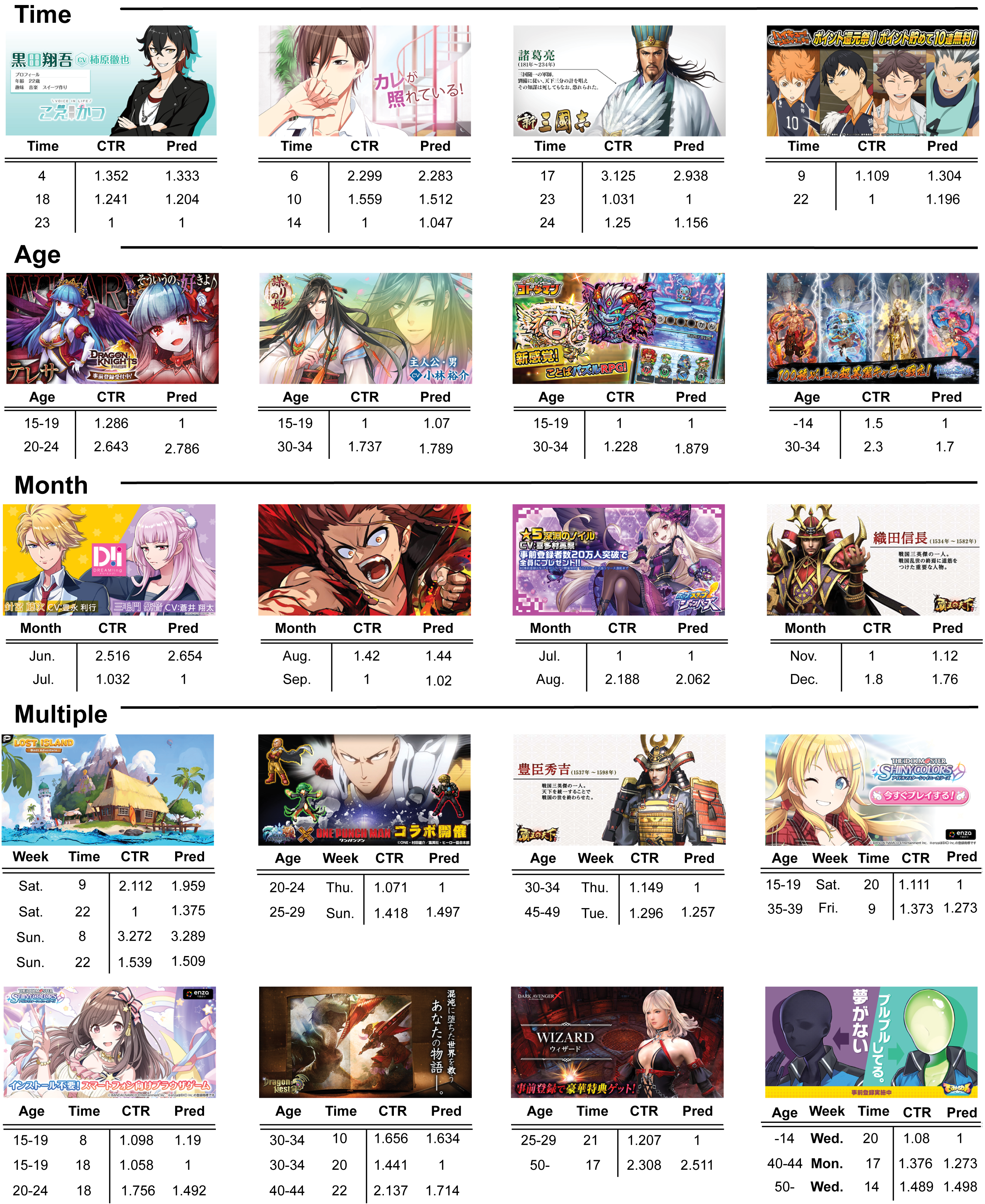}
		\caption{Ad image assessment results of usual cases that follow ad insights of single condition or combination. CTR values are divided by the smallest CTR in each table because of confidentiality issues.}
		\label{results_usual_cases}
	\end{center}
	\vspace{-3mm}
\end{figure*}

\begin{figure*}[]
    \begin{center}
		\includegraphics[width=0.95\textwidth]{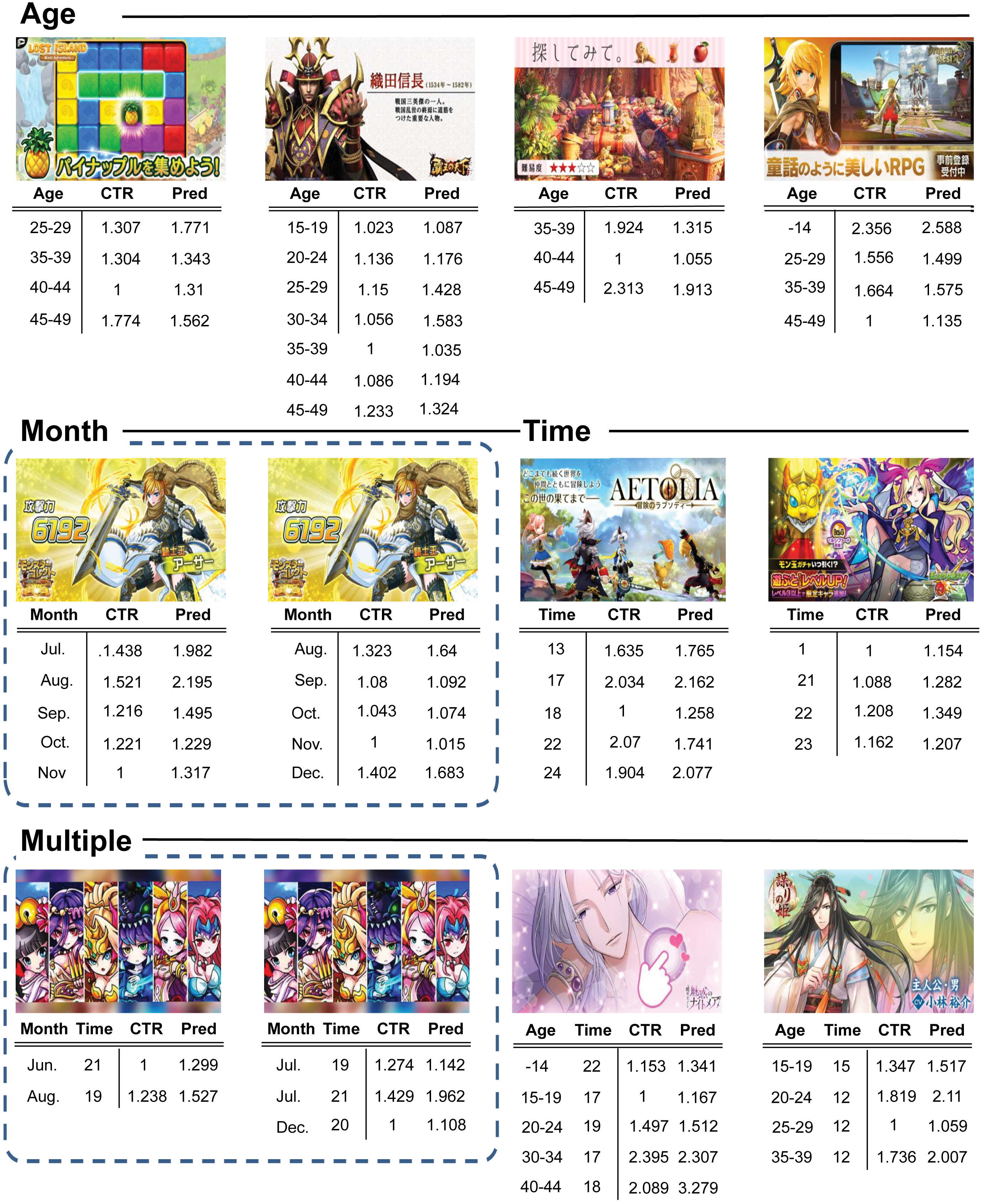}
		\caption{Ad image assessment results of unusual cases that do not follow ad insights of single condition or combination. Dashed rectangle indicates that results inside boxes are of same images with different visual--linguistic auxiliary attributes (title, desc, and OCR). CTR values are divided by the smallest CTR in each table because of confidentiality issues.}
		\label{results_unusual_cases}
	\end{center}
	\vspace{-3mm}
\end{figure*}

Fig.~\ref{results_usual_cases} shows ad image-assessment results for M2FN. Results that have a small difference between the true CTRs and predicted CTRs were selected from the test set. We chose time, age, and month as attributes to determine how this model works according to ad insights. The results are outlined in each row. M2FN is observed to perform better in most cases. 

In contrast, the model is robust even under exceptional circumstances that do not follow the insights. There are unusual cases that counter previous statistical evidence, as in Fig.~\ref{results_unusual_cases}. The results in the first row suggest that CTR increases with age. Other noticeable results are highlighted with dashed rectangles. These are results for identical images but different visual--linguistic auxiliary attributes, such as title, description, and OCR. For example, two ads with identical images may have different catchphrase texts, such as ``this game is a current big hit!'' and ``we payback 50\% of your purchases.'' When the visual--linguistic information displayed to the users is different, the resulting CTRs are also affected. From the results, M2FN is proven capable of handling these unusual cases.

\subsection{Further Analysis}
The results of experiments and statistical analyses provide some insights that can be used to assess the attractiveness of ad images created by human designers. Based on the Real-Ad statistical analyses (ad insights), we reaffirm the results of previous studies, i.e., a single or a combination of auxiliary attributes influences performance. As listed in Table~\ref{table:ablation_aux_real_ad}, four attributes are particularly influential on performance. In addition, with the combined attributes, the fused representation in M2FN fires like an activation function in neurons (e.g., they fire, united but not divided). This would be interesting from the perspective of cognitive science and marketing research. 
\begin{figure*}[t]
    \begin{center}
		\includegraphics[width=0.95\textwidth]{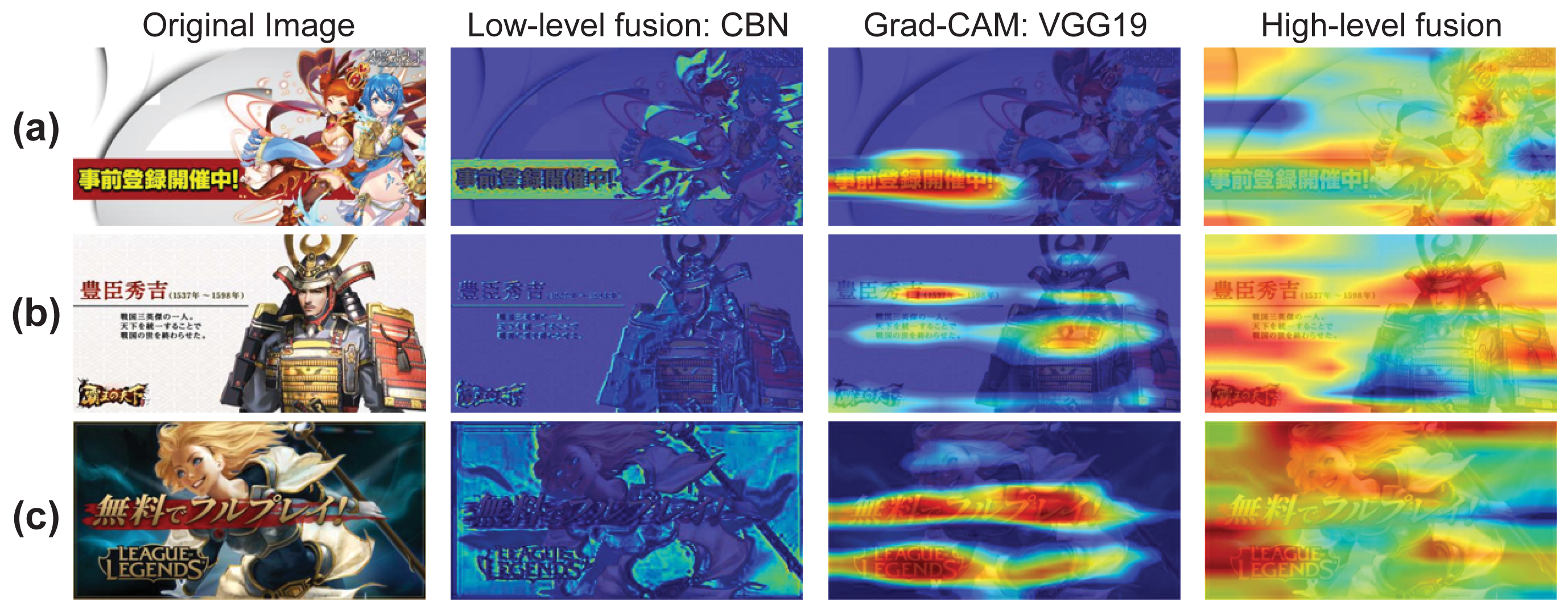}
		\caption{Visualization results using Grad-CAM. Each column displays heatmaps that depict weight vectors of specified layers. Red and blue colors represent highest and the lowest values, respectively. We can find text information such as embedded text and illustrated characters that have significant influence; in the attention layer, they blend. Thus, high-level fusion layer integrates meaningful representations from other modules to assess advertisement images.}
		\label{gradcam_results}
	\end{center}
	\vspace{-3mm}
\end{figure*}

In Fig.~\ref{gradcam_results}, each layer of M2FN is leveraged using Grad-CAM~\cite{selvaraju2017grad} to visualize the operation of each layer in the model. Grad-CAM is a tool that shows which part of an image the neural network sees and how it makes a decision on a particular label. It facilitates the layer in understanding the importance of each neuron with the gradient information. In Fig.~\ref{gradcam_results}, heatmaps are drawn, with red indicating high importance for an area. Interestingly, M2FN seems to be heavily influenced by the text on the image after the extraction of visual features (third column in the figure). Thus, in the attention map (fourth column), a visualization result also shows that partial salience with characters and embedded text is obtained. Based on an analysis of these results, our model demonstrably learns where the visual--linguistic elements that humans consider salient in advertising images are located.  

The visualization results also demonstrate the purpose and structural benefits of multi-step fusion, effectively fusing different levels of ad images by each step, and aiding in in-depth analyses of the ad images.
The model provides a hint regarding the visual--spatial saliency of humans toward the ads. This supports the results of existing literary, cognitive, and marketing science studies. These findings show that M2FN has achieved preliminary success in learning human aesthetic preferences in advertising images.

\begin{figure*}[h]
    \begin{center}
		\includegraphics[width=0.95\textwidth]{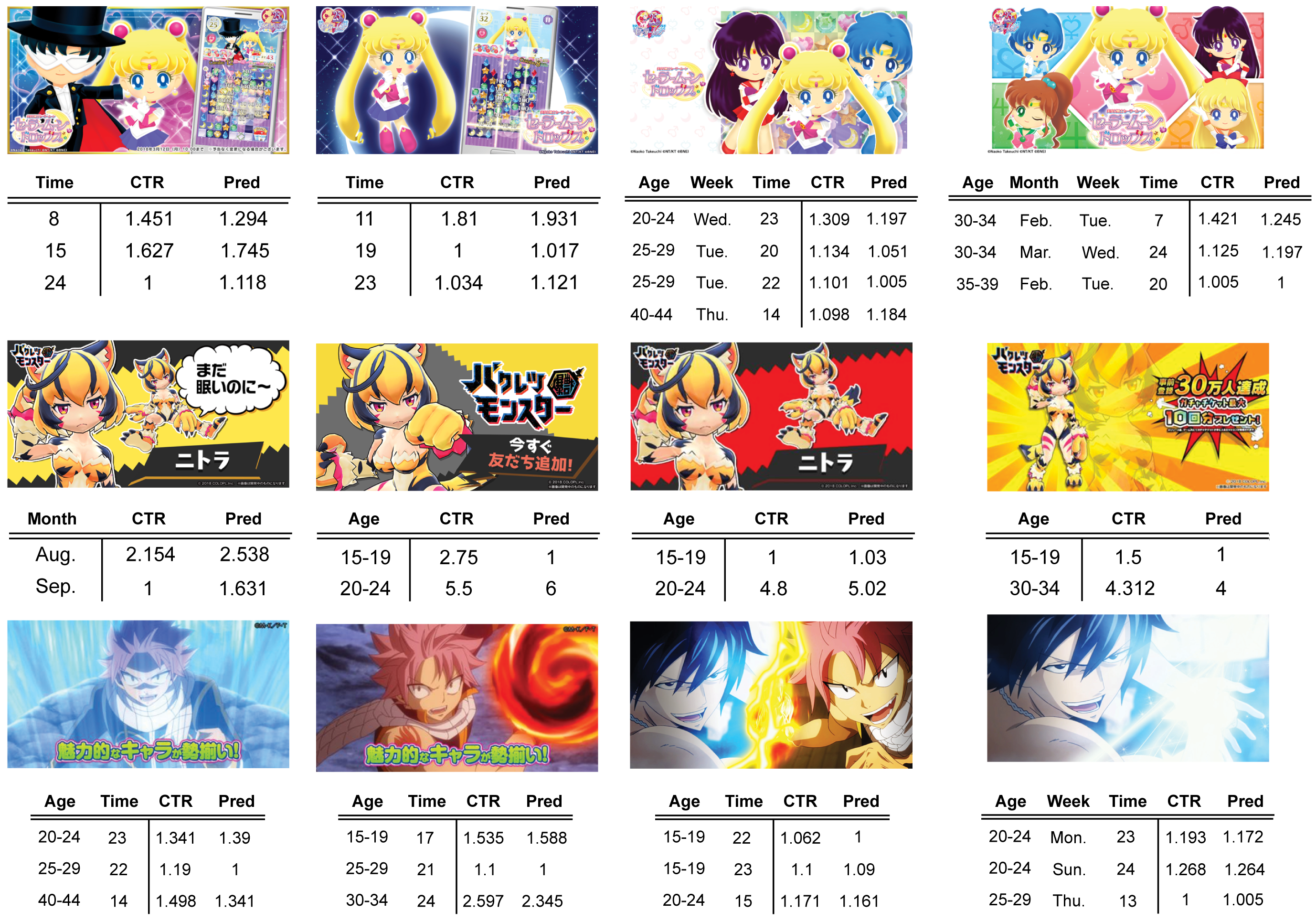}
		\caption{Ad image assessment results of same items and similar images.}
		\label{results_similar_image}
	\end{center}
	\vspace{-3mm}
\end{figure*}

In Fig.~\ref{results_similar_image}, there are several similar ad images for the same item. M2FN works robustly against transformations such as changing the position of the objects or changing colors. As the multi-step structure copes with various conceptual levels, these results indirectly prove that the model performs well with low-level fine-grained controlled images, e.g., changing the angle, font, etc.

Although we achieved excellent performance for the ad image-assessment task, our model is limited to handling only regression problems. It is possible to predict the number of CTRs that would be achieved under a certain condition, but it cannot generate an image that will record a high CTR. Accordingly, we are currently developing a generative model that produces preferable advertising images and a service that provides automated textual advice on the design of ad images and ad displays.

\section{Conclusion}
In this study, we proposed a model for predicting user preference of ad images. The model was tested on an in-house large-scale dataset, containing ad images and auxiliary attributes, called the Real-Ad dataset. We statistically explored the Real-Ad dataset, focusing on the effect of images and auxiliary attributes on human preferences represented as CTR. Inspired by ad insights determined via the analyses, we proposed the novel multi-step modality fusion network (M2FN) model and employed extensive comparison experiments to verify its design. The M2FN effectively integrates ad images and their auxiliary attributes to predict CTR. We evaluated M2FN on the Real-Ad dataset and subsequently validated our method on a benchmark image-assessment dataset and the AVA dataset to verify whether our approach can be applied to conventional image assessment. M2FN achieved better performance on both datasets compared with those of previous SOTA models. With an extensive ablation study, we investigated how each modality fusion works and which auxiliary attributes largely influence user preferences.

\section{Acknowledgments}
The authors thank NAVER AI LAB, NAVER CLOVA for constructive discussion and LINE Corp. for preparing data. We would like to thank Min-Whoo Lee, Chung-Yeon Lee, Taehyeong Kim, Dharani Punithan, Christina Baek, Chris Hickey, Seonil Son, Hwiyeol Jo, and Dong-Sig Han for editing and reviewing this manuscript for English language.

Funding: This research was partly supported by the Institute for Information \& Communications Technology Promotion (2015-0-00310-SW.StarLab, 2017-0-01772-VTT, 2018-0-00622-RMI, 2019-0-01367-BabyMind) and Korea Institute for Advancement Technology (P0006720-GENKO) grant funded by the Korea government. This research was also partly supported by the ICT at Seoul National University.

{\small
\bibliographystyle{ieee_fullname}
\bibliography{egbib}
}

\end{document}